\documentclass[runningheads]{llncs}
\usepackage{graphicx}
\usepackage[]{algorithm2e}
\SetKwInOut{Parameter}{Parameters}
\usepackage{amsmath}
\begin{document}
\title{The data-driven physical-based equations discovery using evolutionary approach}
%
%
\author{Alexander Hvatov \and Mikhail Maslyaev}
\authorrunning{A. Hvatov \and M. Maslyaev}
\titlerunning{Physical-based equations discovery using evolutionary approach}
%
\institute{ITMO University, Kronsersky pr. 49, 197101, St. Petersburg, Russia
\email{alex\_hvatov@itmo.ru}
}
\maketitle              
\begin{abstract}
The modern machine learning methods allow one to obtain the data-driven models in various ways. However, the more complex the model is, the harder it is to interpret. In the paper, we describe the algorithm for the mathematical equations discovery from the given observations data. The algorithm combines genetic programming with the sparse regression.

This algorithm allows obtaining different forms of the resulting models. As an example, it could be used for governing analytical equation discovery as well as for partial differential equations (PDE) discovery.

The main idea is to collect a bag of the building blocks (it may be simple functions or their derivatives of arbitrary order) and consequently take them from the bag to create combinations, which will represent terms of the final equation. The selected terms pass to the evolutionary algorithm, which is used to evolve the selection. The evolutionary steps are combined with the sparse regression to pick only the significant terms. As a result, we obtain a short and interpretable expression that describes the physical process that lies beyond the data.

In the paper, two examples of the algorithm application are described: the PDE discovery for the metocean processes and the function discovery for the acoustics.

\end{abstract}
\section{Introduction}

The modern machine learning methods utilize data-driven models for various purposes. It could be sophisticated surrogate-assisted models \cite{nikitin2019deadline} as well as the complex model identification using evolutionary-based approaches \cite{kovalchuk2018conceptual}.

Nevertheless, the question of the interpretability of the models arises in the applications. Generally, we follow the extensive definition of the model interpretation provided \cite{lipton2016mythos}. Unfortunately, the complexity of the model and interpretability of it, in most cases, require trade-off to obtain good quality and the understanding of how the given model works \cite{molnar2019interpretable}. However, we know the good examples of simple linear regression interpretation \cite{tibshirani1996regression} and inversion of the convolution neural work recognition result \cite{olah2018the}. This kind of interpretation, however, has a drawback, since the interpretability appears only if the obtained result is somehow compared with the human's cognition. 

Physics-based models could be the good examples of the interpretable models \cite{Deep_learning_PDE}. The physics principles provide interpretable basic blocks for the system and mathematics the way of the, possibly, most human-readable form of the model record. However, physical laws are mostly obtained manually by an expert in the field. We could try to derive them automatically in the closed form of the function \cite{schmidt2009distilling}, ordinary differential equation (ODE) \cite{kondrashov2015data}, as well as the partial differential equations (PDE) \cite{berg2019data,schaeffer2017learning}. However, actual realizations require much preliminary work, such as a library of possible terms collection for symbolic regression \cite{rudy2017data}. Such an approach also adds restrictions to the form of the obtained equations mainly because the set of the possible terms is chosen manually \cite{brunton2016discovering}.

In the paper, we propose the method that, in our opinion, allows us to combine the transparency of the physical-based models and flexibility of genetic programming. Moreover, the utility of the sparse regression makes the resulting model form as concise as possible. The method is also similar to the symbolic regression. However, genetic programming allows us to build a flexible library of terms for regression.

On the other hand, such an approach could be considered as an extension to the AutoML methods \cite{guyon2019analysis}. Whereas the last allows obtaining a neural network with the ''best possible'' configuration, the proposed method is used to obtain a model in an extended form. It means that not only the neural network models could be obtained.

The paper is organized as follows in Sec.~\ref{sec:description} the proposed algorithm is described in general. In Sec.~\ref{sec:examples}, two implementation examples are provided: Sec.~\ref{sec:EPDE} contains the example of PDE discovery for the sea surface height data and Sec.~\ref{sec:Floquet} contains the Floquet polynomial discovery for the periodic structure. Sec.~\ref{sec:Conclusion} outlines the main findings of the work and some ideas for future development.

\section{The algorithm description}
\label{sec:description}

Generally, the problem can be summarized as follows: we need to derive a mathematical model (on the current level of framework development, it can be a single equation only) for a physical system. The input is the set of elementary tokens $T = \{t_1, t_2, ..., t_{nt}\}$, where $nt$ is the number of all possible elementary tokens, sufficient for the model creation.

The main goal of the proposed algorithm is the detection of token combinations set (which can be denoted as $C = \{c_m = t_1 \cdot t_2 \cdot \; ... \;  \cdot t_k | \; t_i \in T, m = \overline{1, \; M}\}$ and belongs to the class of all possible token combination sets $\mathbf{C}$, where $k$ is the number number of tokens in the combination and $M$ is the maximum number of terms in the equation), that is able to form the nontrivial linear combination with the minimum absolute value. This approach represents the task to detect structure of function or equation, which can be viewed as the minimization of functional $| \sum_{i=0}^{M} a_i c_i | \rightarrow 0 \; : \; \exists j : a_j \neq 0$, where $c_i$ take roles of the the equation terms, and $a_i$ - weights of the terms. 

\begin{figure*}
\includegraphics[width=1\textwidth]{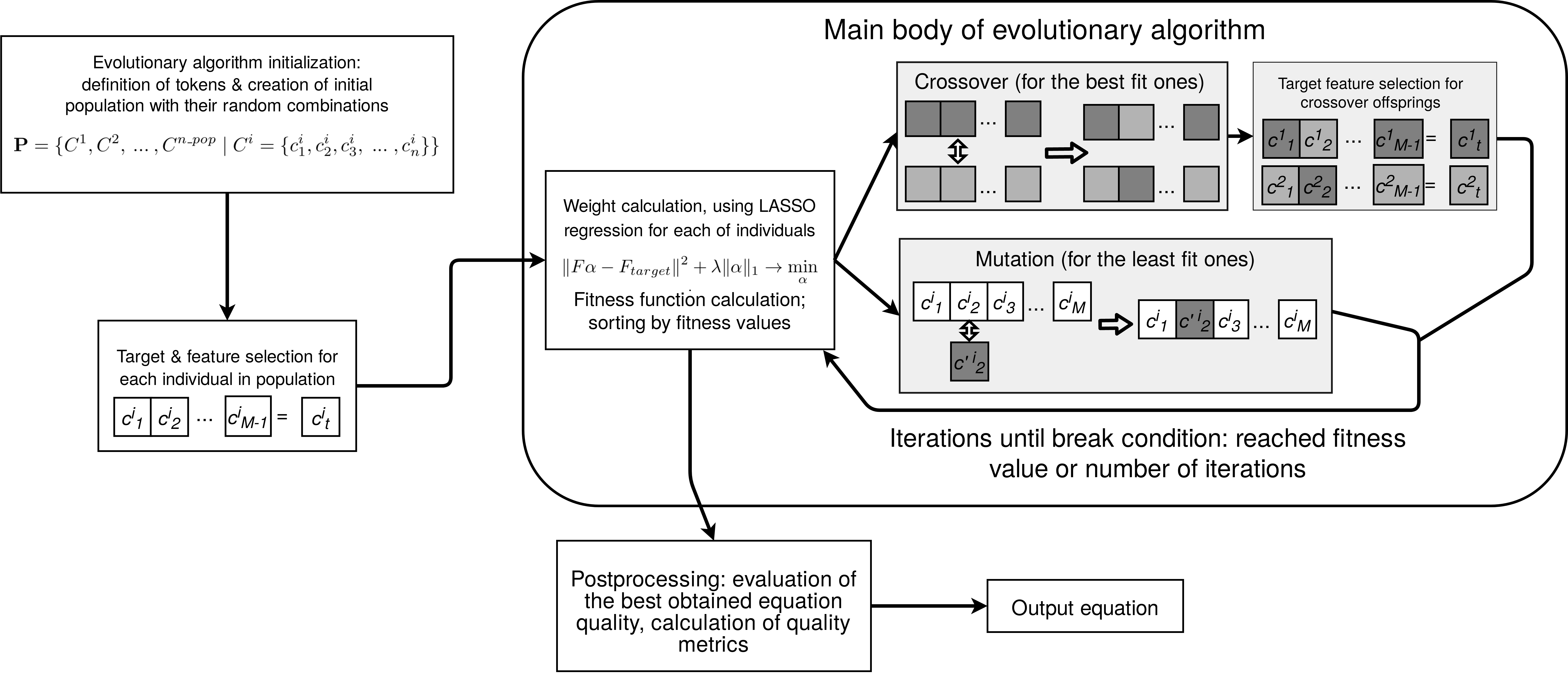}
\caption{The scheme of the method workflow: from the declaration of evolutionary algorithm population to the reception of the resulting equation}
\label{fig:total_scheme}
\end{figure*}

The algorithm consists of three main elements: the building blocks, which we will call tokens below, selection, the evolutionary step, and the sparse regression step. We describe them consequently in this section. The general workflow of the algorithm is presented in Fig.~\ref{fig:total_scheme}.

\subsection{The tokens selection}
\label{sec:tokens}

The tokens could be chosen arbitrary and do not have any restrictions on their nature. However, we stop on the applications of the homogeneous (in terms of the origin) set of the tokens. It means that we take only basic functions or only single derivative terms for evolution.

As an example, it can be all derivatives of the field up to the order $k$. An example of the first derivative token is shown in Eq.~\ref{eq:token_example}. 

\begin{equation}
    \label{eq:token_example}
    c(x,1)=\frac{\partial u}{\partial x}
\end{equation}

As seen token encodes the atomic expression. The form of the expression, as we said above, could be chosen arbitrarily.

From the set of tokens $T$, we compose the words of the length $k$, which is the first hyperparameter of the algorithm. 
We assume that every token in the word has the weighting coefficient and could be replaced with another one without model corruption.

\subsection{The evolutionary part}
\label{sec:evo}


The evolutionary optimization in the algorithm is aimed at the discovery of token combinations set that is able to create the "best" equation. In terms of previously introduced denotation, with this mechanism, we select the set $C$, a linear combination of elements from that can take the least value among its counterparts from the class of sets $\mathbf{C}$. The possibility of extremely high dimensionality of the search space (in usual cases $ ndim = C^{k}_{n \: tokens + k - 1}$; coordinate in each dimension is defined by the weight of terms, corresponding to the dimension, in the final equation) limits the application of other optimization techniques, while the evolutionary methods provide sufficient toolkit for the search.

The initialization of the evolutionary algorithm is done with the creation of the population $\mathbf{P}$ (Eq.~\ref{eq:population}), in that each of the individual solutions will represent a single equation/model. In each of these individuals, $M$ - random token combinations (in a typical case, unordered and with repetitions) are taken as the genes. From each set of combinations, one element is selected as the right part of the equation, while others are assumed to form its left part. Such division excludes the possibility of trivial equation cases, where all of the term weights are equal to 0. 

\begin{equation}
\label{eq:population}
\mathbf{P} = \{C^1, C^2, \: ... \: , C^{n\_pop} \; | \; C^i = \{ c^i_1, c^i_2, c^i_3, \; ... \; , c^i_n \} \}
\end{equation}

For the quality of individual solution evaluation, we shall define the fitness function. The calculation of the fitness function requires the equation's weights: we calculate it as the inverse norm of the difference between the sum of estimated values of token combinations in the left part, and the evaluation of the right part, as shown in Eq.~\ref{eq:fitness_function}, where $\alpha$ stands for the sparse vector of equation weights, $\mathbf{F}_{target}$ is the estimation of the designated as the right part token combination, and $\mathbf{F}$ is comprised of left part token combination. The norm, used in this calculation, is selected according to the specifics of the task. Therefore, the main objective of the algorithm is the detection of such set $C'$, which has the highest possible fitness function value.   

\begin{equation}
\label{eq:fitness_function}
f_{fitness}=\frac{1}{\Vert\mathbf{F} \cdot \mathbf{\alpha}-\mathbf{F}_{target} \Vert}
\end{equation}


The evolutionary search, performed during the algorithm operation, utilizes both mutation and crossover, which are introduced for the alteration of the set of token combinations to create the best one. 

\begin{figure}[ht!]
\includegraphics[width=1\textwidth]{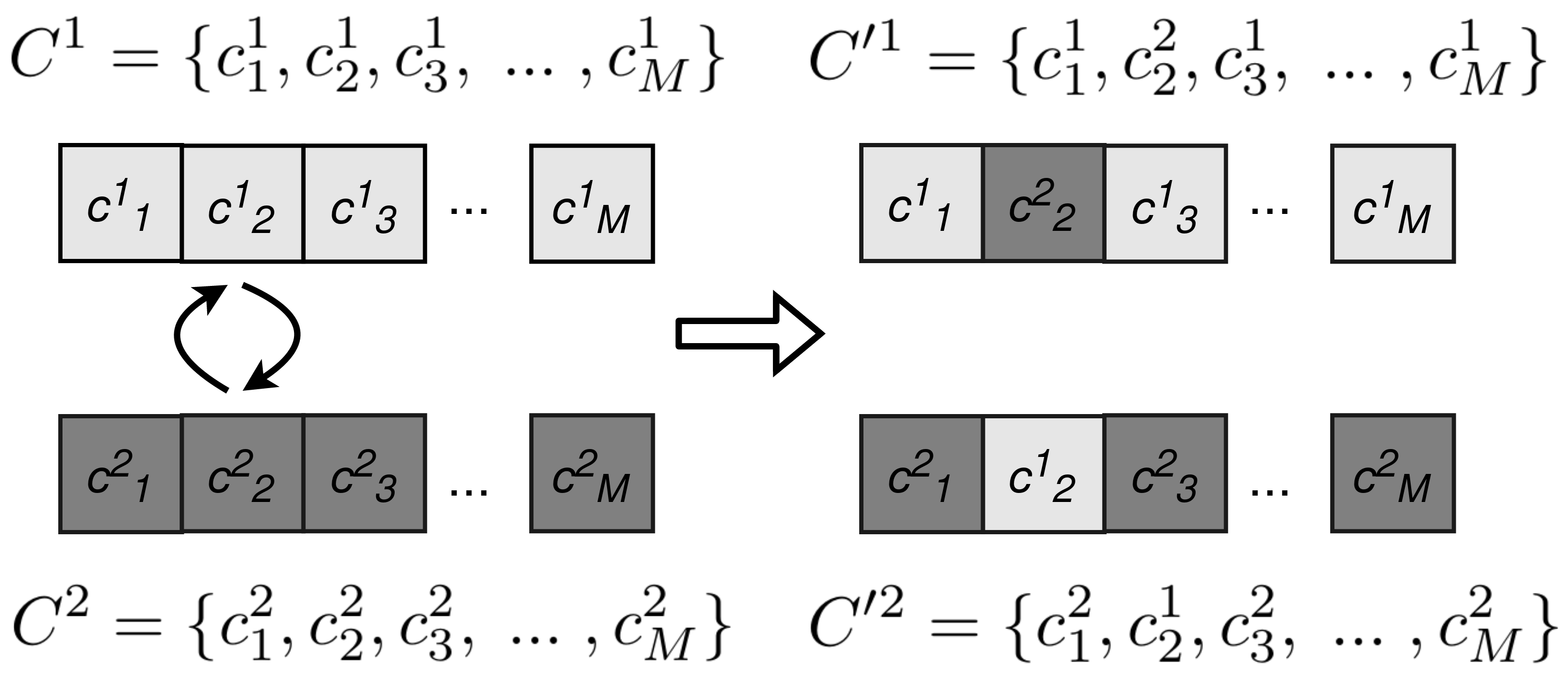}
\caption{Scheme of the crossover}
\label{fig:crossover}
\end{figure}

The crossover operation is defined as the gene exchange between two individuals. In the algorithm, it is represented as the swap of token combination between the models. The scheme of the crossover between two individuals $C^1$ and $C^2$, resulting in offsprings $C'^1$ and $C'^2$, is depicted on Fig~\ref{fig:crossover}. The selection of parents for the procreation is performed via tournament selection, where the appropriate number of tournaments is held to select $a_{proc} \times 100$ \% of the population. Such an approach creates the possibility for the drift of genes from candidates with mediocre fitness values to the next generations. After the offspring creation, the target and the features are also selected in its token sets. The crossover parameter $r_{crossover}$ determines the probability of a single swap during the gene exchange between selected individuals.  

\begin{figure}[ht!]
\includegraphics[width=1\textwidth]{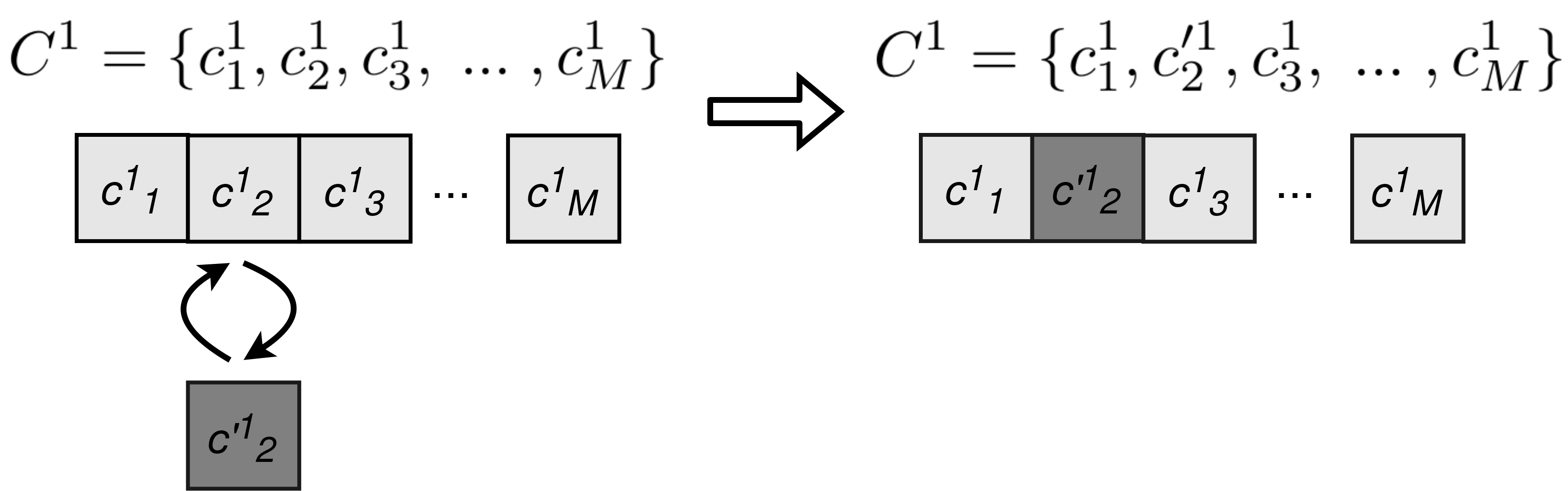}
\caption{Scheme of the mutation, which involves change of gene to the randomly created new one}
\label{fig:mutation_c}
\end{figure}

\begin{figure}[ht!]
\includegraphics[width=1\textwidth]{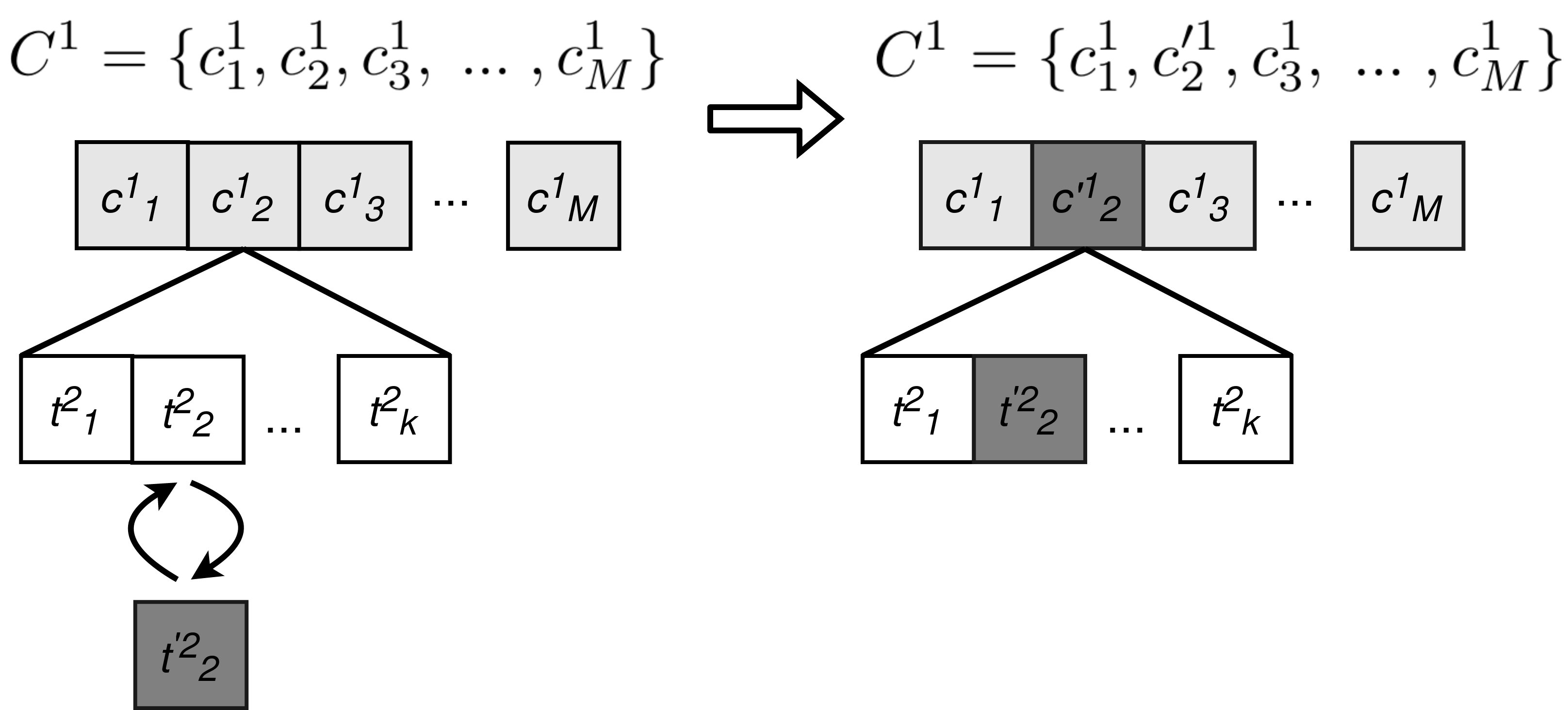}
\caption{Scheme of the mutation, that operated by random change of selected token inside genes}
\label{fig:mutation_t}
\end{figure}

The mutation operator can be defined in two ways: as the possible change of one token in the gene of the existing individual, or as a random swap of the entire gene to the randomly created one. The examples of the implemented mutation operator operations are shown in Fig.~\ref{fig:mutation_c} for the gene swap and Fig.~\ref{fig:mutation_t} for the alteration of a token in the token combination. The type of mutation, applied to the individual is selected randomly in each of the operator application. In order to preserve the currently best individuals, the mutation is forbidden for the specified top percentage of the population $a_{elite}$, according to their fitness value. The mutation rate $r_{mutation}$ defines the probability of each gene in the individual to mutate.  

\subsection{The regression part}
\label{sec:LASSO}

While the previously discussed evolutionary part of the algorithm was developed to discover the best set of token combinations, which will represent the desired structure of the model, the regression methods are utilized to calculate the weights for these terms. Not only the best but also some redundant token combinations may be present in the best discovered set $C'$. Therefore, the task of set filtering is also bestowed to the regression element of the algorithm. The primary method that can perform these jobs is the sparse (regularized) regression, performed with LASSO operator, which is presented in the Eq.~\ref{eq:LASSO}. 

\begin{equation}
\label{eq:LASSO}
    \Vert{F}\alpha -{{F}_{target}}\Vert^{2}+\lambda {\Vert{\alpha }\Vert_{1}}  \to \min \limits_\alpha
\end{equation}

In a typical application, sparse regression involves minimization of the functional, comprised of the sum of the squared L2 norm of the difference between the vector of the target variable and the vector, which is the result of the dot product of the features matrix and the weights vector. While in the majority of cases, the problem can be interpreted in terms of vector operations, sparse regression operator can be extended to the broader class of tasks: for the target, we can use the estimation of the corresponding combination of tokens, and for the features the vector of estimations for other elements in token combinations set. The norm of the difference is also selected according to the type of token representation.

While multiple experiments have proved, that the values of parameters, defining the evolutionary element of the algorithm, have less impact on the resulting equations, alternating mainly the time to achieve the desired solution, the sparsity constant can determine the structure of the final result. Therefore, the task of selecting the optimal value of sparsity constant can be vital for the successful operation of the algorithm. One of the possible solutions to this problem is to initialize the algorithm on the grid of $\lambda$ values and select the best-obtained solution.      

We put the material of the section in a short form of the pseudo-code in Alg.~\ref{alg:pseudo_code}.

\begin{algorithm}
 \KwIn{set of elementary tokens $T$}
\Parameter{$M$ - number of token combinations in a single individual; $k$ - number of elementary tokens in a combination; $n\_pop$ - number of candidate solutions in the population; evolutionary algorithm parameters: number of epochs $n_{epochs}$, mutation $r_{mutation}$ \& crossover rates $r_{crossover}$, part of the population, allowed for procreation $a_{proc}$, number of individuals, refrained from mutation (elitism) $a_{elite}$; sparse regression parameter - sparsity constant $\lambda$}
 \KwResult{set of token combinations $C^{best}$ (if required, with accompanying weights), representing best model/equation for the data}
 Generate population $\mathbf{P}$ of individuals of size $n\_pop$, with $M$ - random permutations of $k$ tokens to form sets $C^j$;
 
 \For{epoch = 1 to $n_{epochs}$}{
  \For{individual in population}{
   Apply sparse regression to individual to calculate weights\;
   Calculate fitness function to individual\;
   }
   Hold tournament selection and crossover\;
   \For{individual in population except $n\_pop \times a_{elite}$ "elite" ones}{
   Mutate individual\;
  }
 }
Select the individual with highest fitness function value as the final solution to the problem\; 
\caption{The pseudo-code of the algorithm operation}
\label{alg:pseudo_code}
\end{algorithm}

\section{Implementation examples}
\label{sec:examples}

In this section, we provide two different examples of algorithm implementation. In Sec.\ref{sec:EPDE} the data-driven PDE discovery is shown in application to the metocean data. In particular, we use sea surface height data obtained from the NEMO ocean model \cite{NEMO_man}. In Sec. \ref{sec:Floquet} the data-driven polynomial Floquet polynomial discovery is considered.  

\subsection{Differential equation discovery for the metocean process}
\label{sec:EPDE}

The data-driven derivation of equations, which describe metocean processes, is one of the suggested applications of the proposed algorithm. The governing equations for the different metocean processes often do not have a closed-form equation. Various parameterizations are used to compensate, for example, the influence of vertical mixing \cite{NEMO_man}. Thus, for the applications, it may be useful to step forward and obtain the data-driven PDE to approximate the process. The approximation could be used either as an advanced parameterization or as a clue to deduce an additional term for the existing equations. 

\subsubsection{Problem statement}

The proposed algorithm is suggested to the class of problems, that involve derivation of the equation for process, that involves variable $f$ and takes place is the specified area $\Omega$ for a period $T$. According to our hypothesis, it can be described with unknown partial differential equation \ref{eq:problem_equation}. The input data for the equation discovery algorithm is presented by sets of measurements $S = \{f_1, f_2, \;... \;, f_n\}$. In the most appropriate and the most common approach, samples are taken on the grid, which can be introduced as $\overline{\gamma} = \{(x_1, x_2, \; ... \; x_k, t) | (x_1, x_2, \; ... \; , x_k) \in \Omega; t \in [0, T] \}$, making sample take form of $f_j = f((x_1^j, x_2^j, \; ... \; x_k^j, t^j))$.

\begin{equation} 
\label{eq:problem_equation}
\begin{cases}
    F(f, \frac{\partial f}{\partial x_1}, \frac{\partial f}{\partial x_2}, ..., \frac{\partial f}{\partial t}, \frac{\partial^2 f}{\partial x_1^2}, \frac{\partial^2 f}{\partial x_2^2}, \; ... \; , \frac{\partial^2 f}{\partial t^2}, \; ... \;) = 0; \\
    G(f) = 0, \; f \in \Gamma(\Omega) \times [0, T] ;
\end{cases}
\end{equation}

The main reason for the introduction of the grid is the simplicity of the derivative calculations. In this case, various robust to the noise in input data methods can be used. The best trade-off between the computational simplicity and resistance to the noise can be achieved with the analytical differentiation of polynomials that are fit over the values in sets of points on the grid. These polynomials are constructed with the least-squares method, finding the coefficients by minimizing the error between the weighted sum of terms and the values in points. Otherwise, in case of non-regular measurements, less conventional methods shall be utilized, such as in situ estimations of the derivatives. 

The tokens for the evolutionary algorithm that will comprise the terms of the resulting equations are composed of various derivatives, taken up to selected order (usually second or third) and along all axis. 

\subsubsection{Experiments}

The conducted experiments were based on the discovery of the governing equation for the dynamics of sea surface height (SSH), obtained from the NEMO Arctic seas model configuration. The datasets included hourly samples on the regular spatial grid on $50 \times 50$ nodes with $5 \; km$ steps for 24 hours intervals. The main dynamics that could be derived with data of this time resolution is connected to the tides, which tend to be oscillations with fixed amplitudes and main periods of roughly 12 hours.   

Preliminary SSH field smoothing is required in the ocean (and, possibly, in the other physical measurement fields) data cases. It is done to reduce the influence of high-frequency noise on derivatives fields. The initial data filtering is performed with a Gaussian smoothing kernel. This results in the improvement of the quality of the derivative fields, that are susceptible to the statistical errors in the initial function field.

The parameters of the evolutionary algorithm during the experiments were set as follows: the crossover $r_{crossover}$ and mutation $r_{mutation}$ rates had values of $0.4$; for procreation, 20\% of the population were selected, and 40\% were excluded from the mutation. The $n\_pop$ - the size of the population was ten individual solutions; each equation could have up to 8 terms altogether in the left and right part, and three tokens can be present in each term.   

\begin{figure}[ht!]
\includegraphics[width=1\textwidth]{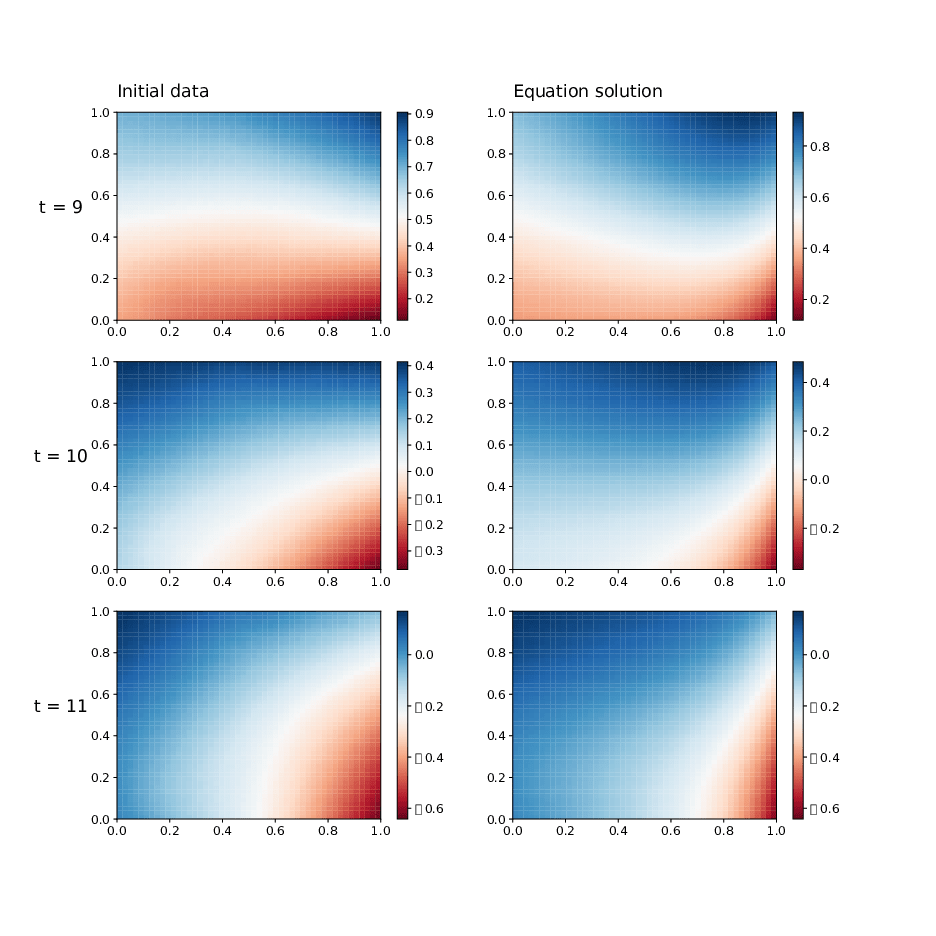}
\caption{Examples of the SSH fields for consequent time frames; initial data and matrix obtained from the equation solution}
\label{fig:Heatmaps}
\end{figure}

In order to validate the performance of the algorithm, the quality of the discovered equations shall be evaluated.  We have selected one day with the governing equation in form Eq.~\ref{eq:discovered_equation}. It was solved, and the results were compared with the initial field. In Fig.~\ref{fig:Heatmaps}, examples of the sea surface height fields, acquired from the data and equation solution, are presented: the left column represents three consequent frames from the framework input data, while their right counterparts are the fields, obtained from the partial differential equation, for the same time steps.

\begin{equation}
\label{eq:discovered_equation}
\frac{\partial f}{\partial x} = -0.05153 \frac{\partial^2 f}{\partial t^2} + 8.508 \frac{\partial^2 f}{\partial x^2} 
\end{equation}

\begin{figure}[ht!]
\includegraphics[width=1\textwidth]{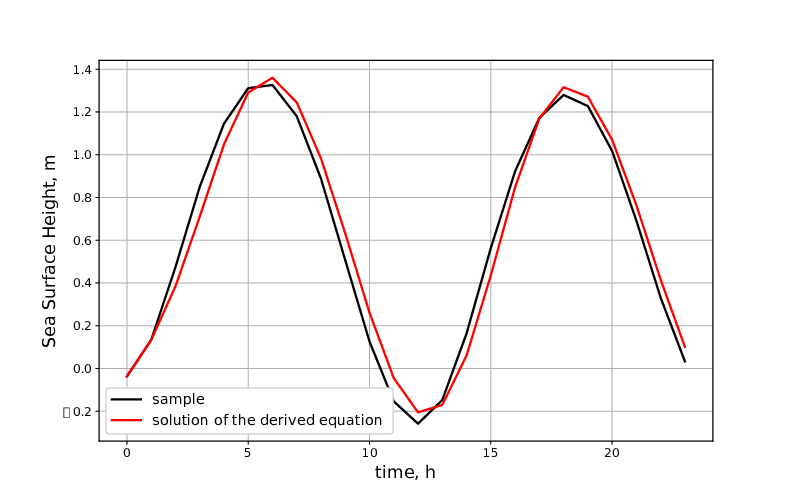}
\caption{Comparison of the SSH field simulation results with the input field for the area center}
\label{fig:graph_25_25}
\end{figure}

The example of a comparison of the initial field and the reconstruction for a single point is presented in Fig.~\ref{fig:graph_25_25}, which represents the time-series, situated in the center of the simulated area. In can be noticed, that the quality of the simulation with the equations is high: the oscillations, that represent tidal dynamics are well preserved, using only the influence of the boundary conditions. It is possible to introduce the metrics of the error, applied to the entire studied area, to prove the effectiveness of the simulation. Root mean square error (RMSE) and mean absolute error (MAE) have values of 0.064 and 0.048 correspondingly. In comparison with the average value of the field (0.625 m), the error values are close to 10\%, which can be relatively good in the tasks of ocean simulation.

\begin{figure}[ht!]
\includegraphics[width=1\textwidth]{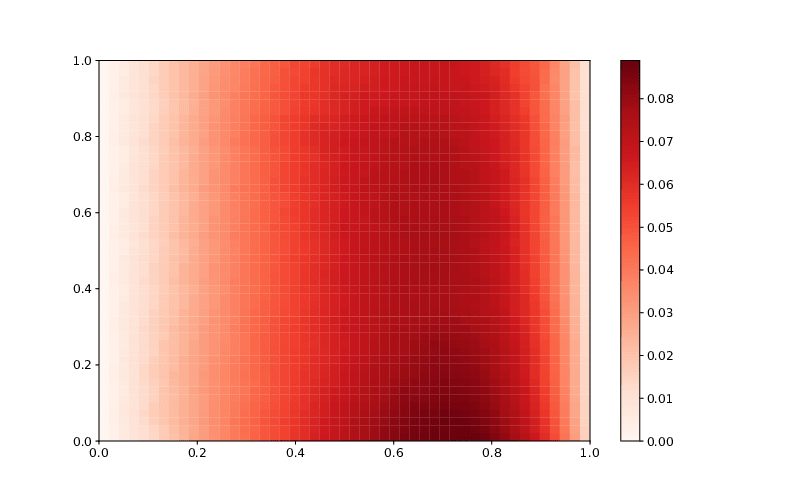}
\caption{Heat map of mean absolute error (MAE) for time-series in each of the grid nodes}
\label{fig:MAE}
\end{figure}

\begin{figure}[ht!]
\includegraphics[width=1\textwidth]{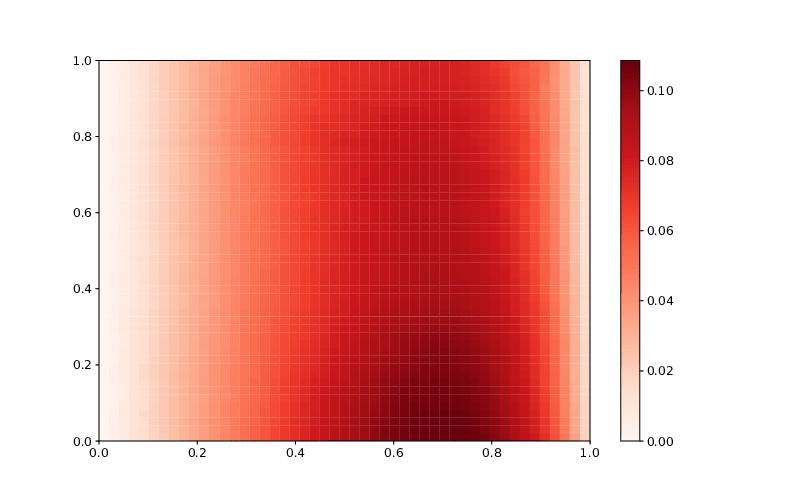}
\caption{Heat map of mean absolute error (MAE) for time-series in each of the grid nodes}
\label{fig:RMSE}
\end{figure}

To understand the spatial distribution of the errors in the field reconstruction, we shall examine the fields of root mean square error (RMSE), presented in Fig.~\ref{fig:RMSE} and mean absolute errors, which is shown in Fig.~\ref{fig:MAE}, calculated over the time-series for each of the grid points. It can be noticed that the error is close to 0 both in the left and right parts of the area. It appears since the boundary conditions for the model were set there. Additionally, the distribution of error is asymmetrical in the central area. Such a difference can be a result of the effect that the amplitudes of tides are higher in the right part (around 2.3), while in the leftmost area, the values are close to 1.0.

\begin{figure}[ht!]
\includegraphics[width=1\textwidth]{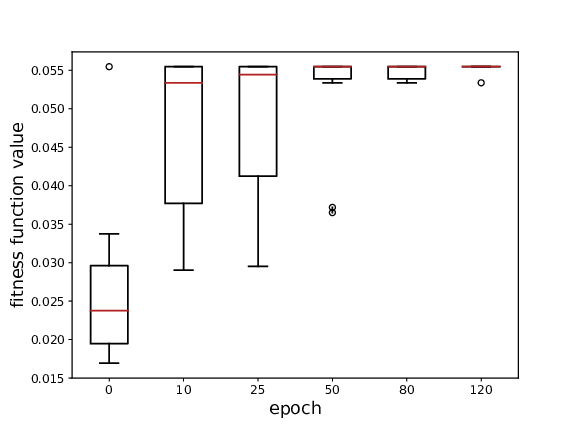}
\caption{Distributions of the best solution fitness values for 10 independent launches of the algorithm}
\label{fig:Fitness_evo}
\end{figure}

In order to analyze the convergence of the algorithm, ten independent runs of the algorithm were held. Results of this experiment are presented in the box plot in Fig.~\ref{fig:Fitness_evo}, where for the specified epoch, the distributions of fitness values of the best candidates are plotted. The majority of the launches have converged to the form of Eq.~\ref{eq:discovered_equation} in approximately 100 epochs. It can be noticed that one of the individuals had the best possible structure since the beginning.  

\subsection{Polynomial discovery for a periodic structure}
\label{sec:Floquet}

Floquet polynomial is an acoustical marker of pass- and stop-bands. The last are zones in a periodic structure, where the waves are propagating free (pass-bands), or the wave propagation is blocked (stop-bands) by the wave interference. Usually, the problem of the zones identification is solved analytically. However, for the complex acoustical models, the analytical solution includes the symbolic determinant computation of order $\geq 100$, which may be computationally expensive.

The simple idea is to replace the determinant computation with the simple numerical algebraic system solution to obtain the data. The obtained data is then used for the polynomial discovery algorithm to approximate the symbolic solution.

\subsubsection{Mathematical problem description}

For this problem, we consider periodic structure, which consists of two periodically repeating blocks of length $1$ and $\gamma$ respectively. The general form of the solution  of the axial rod vibration operator (which is the simple one-dimensional wave equation) is shown in Eq.~\ref{eq:axial_disp}

\begin{equation}
\label{eq:axial_disp}
\begin{array}{cc}
u_i(x)=b_{i,1} \exp{i \Omega x}+b_{i,2} \exp{i \Omega x} (i \, \text{mod} \, 2=1)\\
u_i(x)=b_{i,1} \exp{i \frac{\Omega}{\sigma} x}+b_{i,2} \exp{i \frac{\Omega}{\sigma} x} (i \, \text{mod} \, 2=0)\\
f_i(x)=\frac{d}{dx}u_i(x)
\end{array}
\end{equation}

Variable $\Omega$ is the dimensionless frequency and considered as the problem parameter, and  $\sigma$ is the material difference of two blocks, also considered as the problem parameter. The data for the algorithm is taken from the forcing problem (Eq.~\ref{eq:energy_interf}).

\begin{equation}
\label{eq:energy_interf}
\begin{array}{cc}
u_1(1)=u_2(1)\\
f_1(1)=f_2(1)\\
...\\
u_{n-1}(l_{n-1})=u_n(l_{n-1})\\
f_{n-1}(l_{n-1})=f_n(l_{n-1})\\
f_1(0)=1\\
b_{n,2}=0
\end{array}
\end{equation}

System Eq.~\ref{eq:energy_interf} has a unique solution. After it is found, the solution is used to obtain the Floquet periodicity coefficient approximation in form Eq.~\ref{eq:floquet_approx}.

\begin{equation}
\label{eq:floquet_approx} 
F_n(x,\Omega)=\frac{D_n(x,\Omega)}{D_n(x+(1+\gamma),\Omega)}
\end{equation}

In Eq.~\ref{eq:floquet_approx} with $D_n(x,\Omega)$ displacement of the structure with the unknown constants found from forcing problem Eq.~\ref{eq:energy_interf} is designated.

The solution obtained with the algorithm is compared with the analytical solution, that may be obtained as the determinant of the system Eq.~\ref{eq:axial_periodicity_conditions}

\begin{equation}
\label{eq:axial_periodicity_conditions} 
\begin{array}{cc}
u_1(1)=u_2(1) \\
f_1(1)=f_2(1) \\
u_1(0)=\Lambda u_2(1+\gamma)\\
f_1(0)=\Lambda f_2(1+\gamma)
\end{array}
\end{equation}

Thus, we try to discover Floquet polynomial in form Eq.~\ref{eq:floquet_poly} with algorithm described in Sec.~\ref{sec:description}.  We consider it as the polynomial of degree $2$ in variable $\Lambda$ it could be written as

\begin{equation}
\label{eq:floquet_poly}
D(\Omega)=a_{2}(\Omega)\Lambda^{2}+a_{1}(\Omega)\Lambda+a_{0}(\Omega)=0
\end{equation}

To be brief, main properties of the roots of the Floquet polynomial are stated without any proofs. They could be either complex with property $\text{abs}(\Lambda_i)=1$ (that corresponds to a pass-band) or pure real with property $\text{abs}(\Lambda_j)>1 \; , \; \text{abs}(\Lambda_k)<1 \;, \Lambda_j*\Lambda_k=1$ (stop-band).

For the parameter set $\gamma=1 \;, \sigma=\frac{1}{5}$ the analytical solution has the form Eq.~\ref{eq:axial_rod_floquet}

\begin{equation}
    \label{eq:axial_rod_floquet}
    \Lambda^2+(\frac{169}{60}\cos{(6 \Omega)}-\frac{49}{60}\cos{(4\Omega)})\Lambda+1
\end{equation}

\subsubsection{Algorithm quality assessment}

After analytical solution is found, the next step is to compare it with the one, discovered by the algorithm. Following Sec.~\ref{sec:description}, the set of the tokens is defined first.
From form of the analytical solution Eq.~\ref{eq:axial_rod_floquet}, we deduce that tokens for this case have the form Eq.~\ref{eq:tokens_floquet}.

\begin{equation}
    \label{eq:tokens_floquet}
    T_j(A,B)=(A \cos(B \Omega)) \Lambda^j
\end{equation}

Token in the equation has three parameters $A, B, j$, we evolve parameter $B$ and allow to appear new tokens in the polynomial during the evolution. Since we want only to illustrate the approach, we allow only the degrees $j=0,1,2$ to appear. The sparse regression is done with respect to the parameter $A$. 

Second step is to define evolutionary operators. The evolution is done in the same way as in the Sec.~\ref{sec:EPDE}. The mutation allows a new term to appear. However, the maximal number of terms in the sum is restricted. The multiplication of the terms is forbidden sine we know the resulting form of the analytical solution. The algorithm's convergence rate differs insignificantly from the Sec.~\ref{sec:EPDE}. Thus, we do not stop on the evolutionary part and show only the quality of the result compared to the analytical solution.

\begin{figure}[ht!]
\includegraphics[width=1\textwidth]{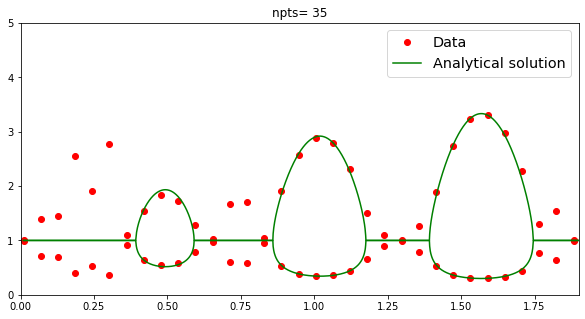}
\caption{The data (dots) and the analythical solution for $n=35$ data points}
\label{fig:floquet_data}
\end{figure}

An example of data in Fig.~\ref{fig:floquet_data} shown $n=35$ uniformly taken points, found from Eq.~\ref{eq:floquet_approx} in the range $\Omega=[0,2]$.

\begin{figure}[ht!]
\includegraphics[width=1\textwidth]{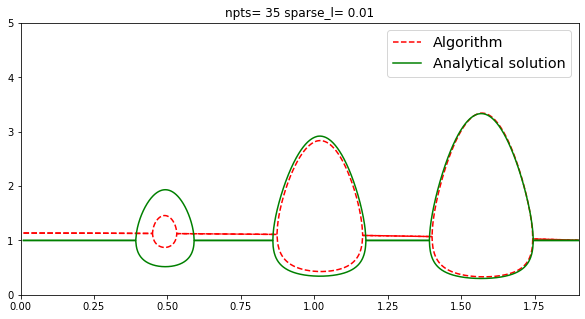}
\caption{The discovered polynomial (dashed) and the analythical solution for $n=35$ data points}
\label{fig:single_floquet_poly}
\end{figure}

Straight lines in analythical solution in  Fig.~\ref{fig:floquet_data} corresponds to a pass-bands and vice versa eliptical curves are stop-bands. The discovered polynomial for the data, shown in  Fig.~\ref{fig:floquet_data} is shown in Fig.~\ref{fig:single_floquet_poly}.

It is seen, that discovered polynomial can filter out the data and show approximately the picture of the stop- and pass-bands. It means that the algorithm is able to reproduce this physical process correctly.

\begin{figure}[ht!]
\includegraphics[width=1\textwidth]{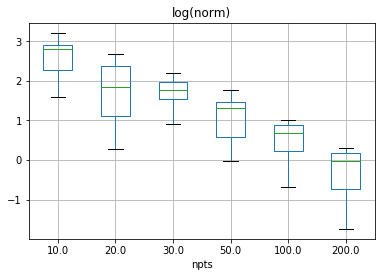}
\caption{Distribution of the $\log ({\text{RMSE}})$  for the one hundred consecutive runs with a different number $npts$ of data points}
\label{fig:floquet_distr}
\end{figure}

The quality of the algorithm is measured as the RMSE between roots of the polynomial Eq.~\ref{eq:axial_rod_floquet} and the resulting one at the points in the range $\Omega=[0,2]$ with discrete steps taken with $\Delta=0.001$. Distribution of the $\log ({\text{RMSE}})$ for the one hundred consecutive runs with a different number $npts$ of data points are shown in Fig.~\ref{fig:floquet_distr}.

It is seen that the algorithm also converges to an analytical solution with an increase in the data points amount. This result gives the reason to believe that in more complex cases, i.e., when the operator is more complex, the algorithm may also give an approximation to the stop- and pass-bands picture.

\section{Conclusion}
\label{sec:Conclusion}

In the paper, we describe the algorithm for the physical-based equations discovery. We want to outline the following properties of it:

\begin{enumerate}
    \item[*] It does not depend on the form of the equation: it could be a polynomial, differential equation, and potentially the other models. However, additional work for the adaptation for each type of the equation is required;
    \item[*] The genetic programming can be used to obtain an optimal bag of the terms from the small set of the building blocks and preliminary defined mutation and crossover rules;
    \item[*] The sparse regression step allows one to filter out the non-descriptive terms that lead to a robust model. As an additional advantage the resulting model has the short form of the expression, which makes the interpretation process easier;
    \item[*] PDE discovery implementation is noise stable even for multi-dimensional data cases. The overall performance of the algorithm implementation allows reproducing tempo-spatial physic fields correctly.
\end{enumerate}

In the future, we plan to combine different types of terms in the co-evolution step that will make it possible to discover models like non-homogeneous PDE. The second development direction is the systems of the generation of the expression.

\section*{Acknowledgments}

This research is financially supported by The Russian Scientific Foundation, Agreement \#19-11-00326.

%
%
%
 \bibliographystyle{splncs04}
\bibliography{references}

\end{document}